\pdfoutput=1

\documentclass[11pt]{article}

\usepackage[final]{naacl2021}

\usepackage{times}
\usepackage{latexsym}
\usepackage{amsmath}
\usepackage{amsfonts}

\usepackage[T1]{fontenc}

\usepackage[utf8]{inputenc}

\usepackage{microtype}

\usepackage{graphicx}
\usepackage{booktabs}
\usepackage{multirow}
\usepackage{array}
\usepackage{color}

\title{Improving Cross-Modal Alignment in Vision Language Navigation via Syntactic Information}

   \author{Jialu Li \quad \quad Hao Tan \quad \quad Mohit Bansal
 \\ 
   UNC Chapel Hill\\ 
   \texttt{\{jialuli, airsplay, mbansal\}@cs.unc.edu}
}

\begin{document}
\maketitle
\begin{abstract}
Vision language navigation is the task that requires an agent to navigate through a 3D environment based on natural language instructions. One key challenge in this task is to ground instructions with the current visual information that the agent perceives. Most of the existing work employs soft attention over individual words to locate the instruction required for the next action. However, different words have different functions in a sentence 
(e.g., modifiers convey attributes, verbs convey actions). 
Syntax information like dependencies and phrase structures can aid the agent to locate important parts of the instruction. 
Hence, in this paper, we propose a navigation agent that utilizes syntax information derived from a dependency tree to enhance alignment between the instruction and the current visual scenes. 
Empirically, our agent outperforms the baseline model that does not use syntax information on the Room-to-Room dataset, especially in the unseen environment. Besides, our agent achieves the new state-of-the-art on Room-Across-Room dataset, which contains instructions in 3 languages (English, Hindi, and Telugu). 
We also show that our agent is better at aligning instructions with the current visual information via qualitative visualizations.\footnote{Code and models: \url{https://github.com/jialuli-luka/SyntaxVLN}}

\end{abstract}

\section{Introduction}
Vision-Language Navigation defines the task of requiring an agent to navigate through a visual environment based on natural language instructions \citep{anderson2018vision, Misra2018MappingIT, chen2019touchdown, jain-etal-2019-stay, nguyen-daume-iii-2019-help, thomason2020vision}. This task poses several challenges. To complete this task, an embodied agent needs to perceive the surrounding environment, understand the given natural language instructions, and most importantly, ground (or align) the instruction in the visual scenes. In this paper, we aim to make one step towards grounding
natural language instructions with visual environment via syntax-enriched alignment. 

\begin{figure}[t]
\centering
\includegraphics[width=1.0\columnwidth]{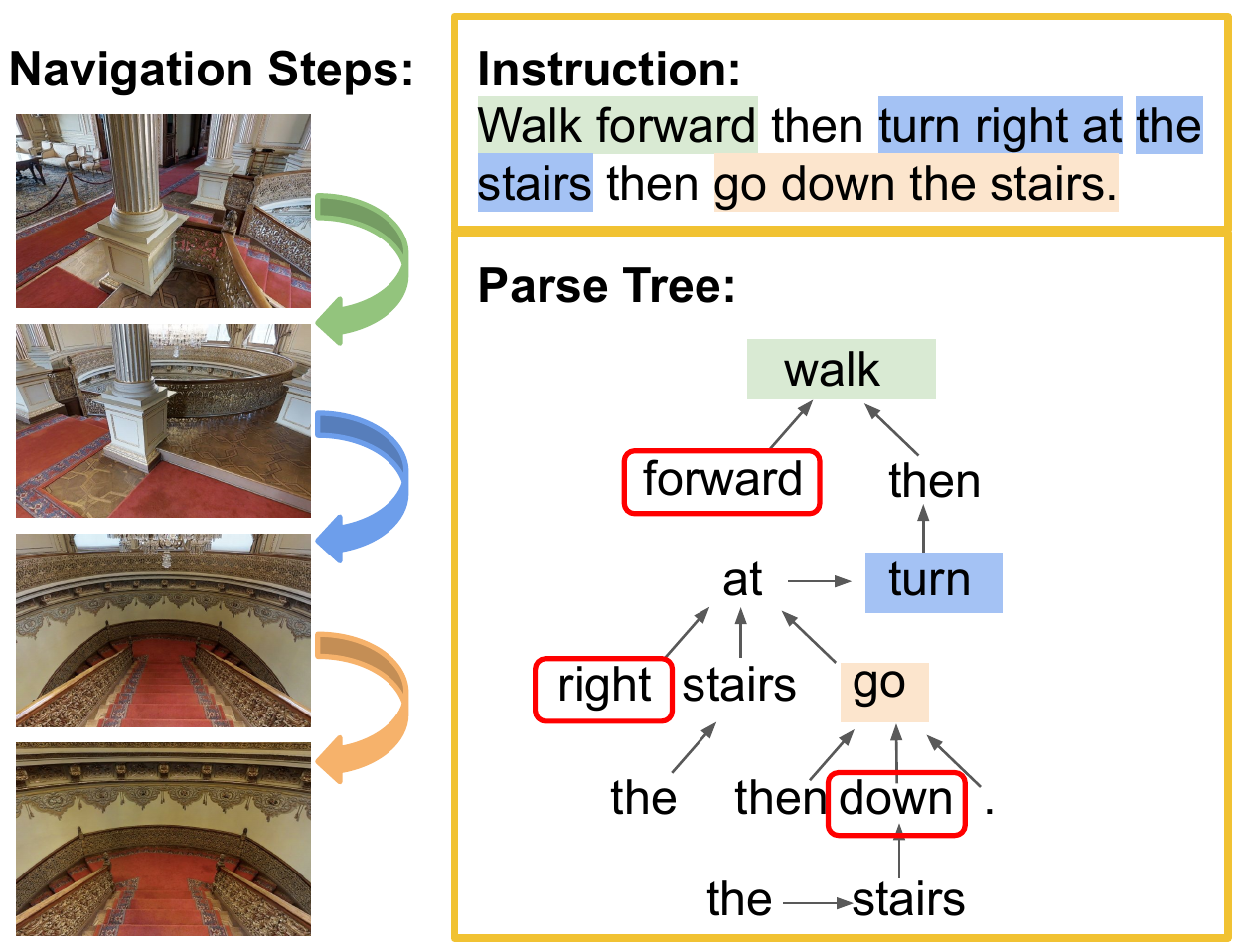}
\caption{An example in the Room-to-Room task. We generate the dependency parse tree for the instruction. 
The words are grouped by the head node (highlighted in the tree). Each sub-instruction (i.e., grouped words) corresponds to one step in the navigation with the same color. Modifiers in red boxes can be easily identified from the tree structure. 
}
\label{fig1}
\end{figure}

Recently, several approaches were proposed to solve the Vision-Language Navigation task with better interactions between natural language instructions and visual scenes \citep{fried2018speaker, wang2019reinforced, landi2019perceive, wang2020active,  Huang2019MultimodalDM, hu-etal-2019-looking, majumdar2020improving, ma2019selfmonitoring, qi2020object, zhu2020vision, Zhu_2020_CVPR}. Some approaches utilize soft attention over individual words for better cross-modal grounding, while others improve co-grounding with better language and vision representation and additional alignment module. 

Although these models achieve significant improvement in performance, they do not explicitly consider syntactic linguistic information in their alignment and decision-making. 
We argue that the syntactic information (e.g., phrases, word functions, modifiers) captured by dependency parse trees is crucial for accurate alignment between the instructions and the environment. 
As shown in Figure \ref{fig1}, for the instruction \textit{``Walk forward then turn right at the stairs then go down the stairs."}, the dependency parse tree effectively aggregates syntactically-close words together for the agent (e.g., groups phrase information \textit{``Walk forward"} at node \textit{``Walk''}), and each phrase here corresponds to one navigation action. Besides, the dependency tree structure also helps identify modifiers like \textit{``forward''} and \textit{``right''}. This syntactic information helps the agent identify important words, locate phrases (e.g., sub-instructions), and learn a better alignment between the instruction and the visual environment.

Therefore, in this paper, we propose an encoder module that can incorporate simple but important syntactic information from parse trees for the vision-language-navigation task.
Our proposed encoder utilizes the ChildSum Tree-LSTM \citep{tai-etal-2015-improved} over a dependency tree to achieve a syntax-aware representation of the instruction, enabling the agent to focus on more syntactically important words and align not only words but also phrases with the visual scenes.

We conduct experiments on both the Room-to-Room (R2R) dataset \citep{anderson2018vision} and the Room-across-Room (RxR) dataset \citep{ku2020room}. Empirical results show that our proposed approach significantly improves the performance over the baseline model on success rate 
and achieves the new state-of-the-art (at the time of submission) 
on the RxR dataset, which contains instruction in three languages (English, Hindi, and Telugu). Moreover, by using structured information from syntax, we are also able to avoid word-level shallow overfitting of the model and hence achieve better generalization in the unseen environment. 
Our analysis further shows that our syntax-aware agent has better interpretability and learns better cross-modality matching. 

\section{Related Work}

\paragraph{Visual and Textual Grounding in VLN.}
In vision-language navigation tasks, visual and textual co-grounding aims to learn the relationship between natural language instructions and the visual environments. 
A main line of research in VLN utilizes soft attention over individual words for cross-modal grounding in both the natural language instruction and the visual scene \citep{wang2018look, wang2019reinforced, tan-etal-2019-learning, landi2019perceive, xia2020multi, wang2020soft, wang2020active, xiang2020learning, DBLP:conf/acl/ZhuHCDJIS20}. Other works improve vision and language representations \citep{hu-etal-2019-looking, Li2019RobustNW, huang2019transferable, Huang2019MultimodalDM, hao2020towards, majumdar2020improving} and propose an additional progress monitor module \citep{ma2019regretful, ma2019selfmonitoring, ke2019tactical} and object and action aware modules \citep{qi2020object} that aid co-grounding. 

The closest work to ours is from \citet{hong2020sub}, where they use the dependency tree to generate pre-divided sub-instructions, and then propose a shifting module to select and attend to a single sub-instruction at each time-step, which implicitly captures some syntax information. Compared with them, we directly use the dependency tree to explicitly incorporate syntactic information and get syntax-aware instruction representations, hence achieving substantial improvement in generalizing to the unseen environment.

\paragraph{Tree-based Language Representations.}
Dependency tree provides essential syntactic information for understanding a sentence.
Tree-LSTM \citep{tai-etal-2015-improved} has been widely used to encode parsed tree information and shown improvement over multiple tasks, such as relation extraction \citep{miwa-bansal-2016-end, geng2020semantic}, machine translation \citep{su2020neural, choi2017learning, eriguchi-etal-2016-tree}, dialogue \citep{rao2019tree}, and language inference \citep{chen-etal-2017-enhanced}. We are novel in incorporating a dependency tree into the vision-language navigation task via a Tree-LSTM for better phrase-level alignment between the visual environment and language instructions.

\section{Method}
\begin{figure}[t]
\centering
\includegraphics[width=1\columnwidth]{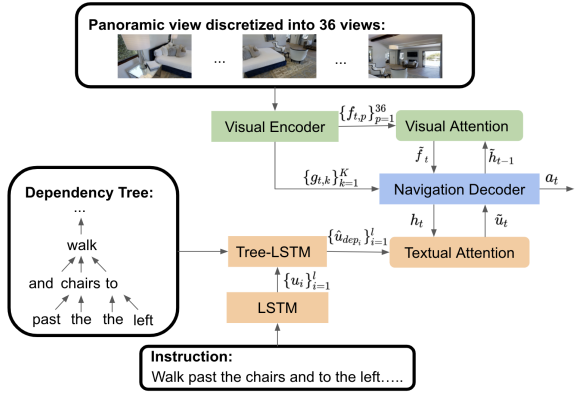}
\caption{Architecture for our syntax-aware agent. 
}
\label{fig2}
\end{figure}

As illustrated in Figure \ref{fig2}, our base model follows the sequence-to-sequence architecture of previous VLN agents. Our tree-based encoder module is built on top of the strong Environment Drop Agent \citep{tan-etal-2019-learning}. The main difference is that we employ a tree-based language encoder to encode dependency tree information to allow better language grounding. At each time step, we ground all the encoded nodes (i.e., syntax-aware representations) in the dependency tree with the visual information to get the attended textual representation. 

\paragraph{Generating Dependency Tree Representation with Tree-LSTM.}
We first generate the dependency parse tree with Stanford CoreNLP \citep{manning2014stanford} for English instructions and Stanza Toolkit \citep{qi-etal-2020-stanza} for Hindi and Telugu instructions. Since the tree-structure is invariant to the children's order (i.e., switching the order of the children of a node doesn't change the tree structure), directly using a Tree-LSTM over an embedding layer may lose important word order information in the instruction. Thus, here we use a bidirectional LSTM with an embedding layer to generate word representations that preserve sequential information of the instruction. Specifically, given an instruction $\{w_i\}_{i=1}^{l}$, we generate syntax-aware representation $\{\hat{u}_{dep_i}\}_{i=1}^{l}$ as:
\begin{align}
    \hat{w_i} &= \mathrm{Embedding}(w_i) \\
    u_1, u_2, ..., u_l &= \mathrm{Bi\mbox{-}LSTM}(\hat{w}_1, \hat{w}_2, ..., \hat{w}_l) \\
    \{\hat{u}_{dep_i}\}_{i=1}^{l} &= \mathrm{Tree\mbox{-}LSTM}(\{u_i\}_{i=1}^{l})
\end{align}

\paragraph{Visual Encoder and Navigation Decoder.} 
Given panoramic features $\{f_{t,p}\}_{p=1}^{36}$ and visual representation $\{g_{t,k}\}_{k=1}^K$ for $K$ navigable locations at time step $t$ \footnote{Details for generating these features are in Appendix.}, we picks the next viewpoint from $K$ navigable locations as:
\begin{align}
    p_t(a_t = k) &= \mathrm{Softmax}_k(g_{t,k}^TW_G\widetilde{h}_t)
\end{align}
where $\widetilde{h}_{t}$ is the context aware hidden states, and $W_G$ is learned weight parameter. Specifically, we compute the $\widetilde{h}_{t}$ as:
\begin{align}
    \beta_{t,p} &= \mathrm{softmax}_p(f_{t,p}^TW_F\widetilde{h}_{t-1})\\
    \widetilde{f}_t &= \sum_{p}{\beta_{t,p}f_{t,p}}\\
    h_t &= \mathrm{LSTM}([\widetilde{f}_t; \widetilde{a}_{t-1}], \widetilde{h}_{t-1})\\
    \gamma_{t,i} &= \mathrm{softmax}_i(\hat{u}_{dep_i}^TW_Uh_t)\\
    \widetilde{u}_t &= \sum_i{\gamma_{t,i}\hat{u}_{dep_i}}\\
    \widetilde{h}_t &= \tanh(W_M[\widetilde{u}_t;h_t])
\end{align}
where $\widetilde{a}_{t-1}$ is the previous action embedding, $\widetilde{f}_t$ is the attended panoramic representation, and $h_t$ is the decoder hidden state. $W_F$, $W_U$, $W_M$ are learned weight parameters. We compute the attended language representation over all dependency node representations which are aware of syntax information. 

We use a mixture of imitation learning and reinforcement learning to train the agent. Details can be found in Appendix.

\begin{table*}[]
\begin{small}
\centering
\begin{tabular}{p{0.3\columnwidth}>{\centering\arraybackslash}p{0.1\columnwidth}>{\centering\arraybackslash}p{0.1\columnwidth}>{\centering\arraybackslash}p{0.1\columnwidth}>{\centering\arraybackslash}p{0.1\columnwidth}>{\centering\arraybackslash}p{0.1\columnwidth}>{\centering\arraybackslash}p{0.1\columnwidth}>{\centering\arraybackslash}p{0.1\columnwidth}>{\centering\arraybackslash}p{0.1\columnwidth}>{\centering\arraybackslash}p{0.1\columnwidth}>{\centering\arraybackslash}p{0.1\columnwidth}}
\hline 
\multicolumn{1}{c}{\textbf{Models}} & \multicolumn{5}{c}{\textbf{Val Seen}} & \multicolumn{5}{c}{\textbf{Val Unseen}} \\ \hline
& SR(\%) & \ SPL & nDTW & sDTW & CLS & SR(\%) &\  SPL & nDTW & sDTW & CLS \\ \hline
\textbf{EnvDrop} & 58.3 & 0.55 & 0.67 & 0.52 & 0.67 & 45.3 & 0.42 & 0.58  & 0.39 & 0.58  \\ 
\textbf{+LSTM} & 60.3& 0.57& 0.69& 0.55&0.69 & 46.4& 0.43& 0.58& 0.40&0.58  \\
\textbf{+syntax} & \textbf{62.6} & \textbf{0.60} & \textbf{0.70} & \textbf{0.56} & \textbf{0.70} & \textbf{49.0} & \textbf{0.45} & \textbf{0.59} & \textbf{0.42} & \textbf{0.59} \\  \hline
\end{tabular}
\caption{Comparison of the model with and without our tree-based encoder on the seen validation set and unseen validation set of R2R dataset. 
}
\label{table2}
\end{small}
\end{table*}

\begin{table*}[]
\begin{small}
\centering
\begin{tabular}{p{0.3\columnwidth}>{\centering\arraybackslash}p{0.1\columnwidth}>{\centering\arraybackslash}p{0.1\columnwidth}>{\centering\arraybackslash}p{0.1\columnwidth}>{\centering\arraybackslash}p{0.1\columnwidth}>{\centering\arraybackslash}p{0.1\columnwidth}>{\centering\arraybackslash}p{0.1\columnwidth}>{\centering\arraybackslash}p{0.1\columnwidth}>{\centering\arraybackslash}p{0.1\columnwidth}>{\centering\arraybackslash}p{0.1\columnwidth}>{\centering\arraybackslash}p{0.1\columnwidth}}
\hline 
\multicolumn{1}{c}{\textbf{Models}} & \multicolumn{5}{c}{\textbf{Val Seen}} & \multicolumn{5}{c}{\textbf{Val Unseen}} \\ \hline
& SR(\%) & \ SPL & nDTW & sDTW & CLS & SR(\%) &\  SPL & nDTW & sDTW & CLS \\ \hline
\textbf{EnvDrop (en)} &48.1&\textbf{0.44}&0.57&0.40&0.61&38.5&0.34&0.51&0.32 & 0.54 \\
\textbf{+syntax (en)} &\textbf{48.1}&0.44&\textbf{0.58}&\textbf{0.40}&\textbf{0.61}&\textbf{39.2}&\textbf{0.35}&\textbf{0.52}&\textbf{0.32}&\textbf{0.56} \\
\textbf{EnvDrop (hi)} &49.6&0.45&0.57&0.41&0.61&39.9&0.35&0.49&0.32&0.53 \\ 
\textbf{+syntax (hi)} &\textbf{55.2}&\textbf{0.52}&\textbf{0.61}&\textbf{0.46}&\textbf{0.64}&\textbf{42.5}&\textbf{0.38}&\textbf{0.54}&\textbf{0.35}&\textbf{0.58} \\
\textbf{EnvDrop (te)} &45.8&0.42&0.56&0.38&0.60&38.3&0.34&0.50&0.31&0.54\\ 
\textbf{+syntax (te)} &\textbf{49.1}&\textbf{0.46}&\textbf{0.59}&\textbf{0.41}&\textbf{0.63}&\textbf{38.4}&\textbf{0.35}&\textbf{0.52}&\textbf{0.32}&\textbf{0.56}\\ 
\hline
\end{tabular}
\caption{Comparison of the model with the tree-based encoder and without the tree-based encoder on the seen validation set and unseen validation set of RxR dataset.}
\label{table3}
\end{small}
\end{table*}

\begin{figure*}[t]
\centering
\includegraphics[width=1.9\columnwidth]{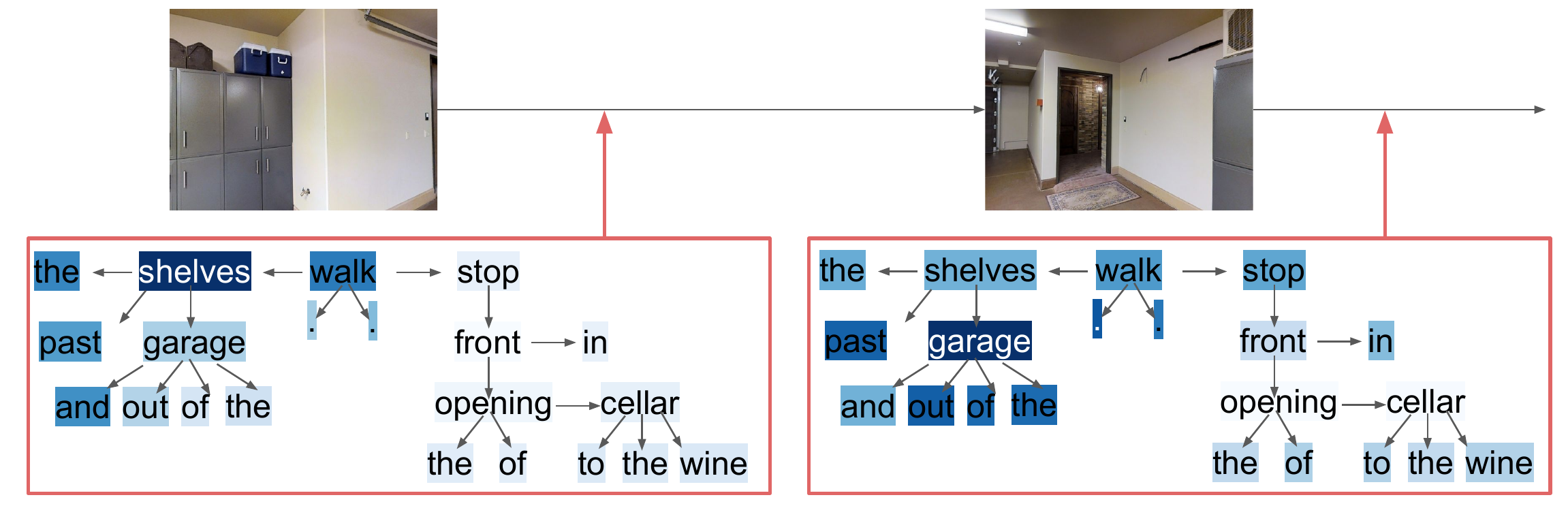}
\includegraphics[width=1.9\columnwidth]{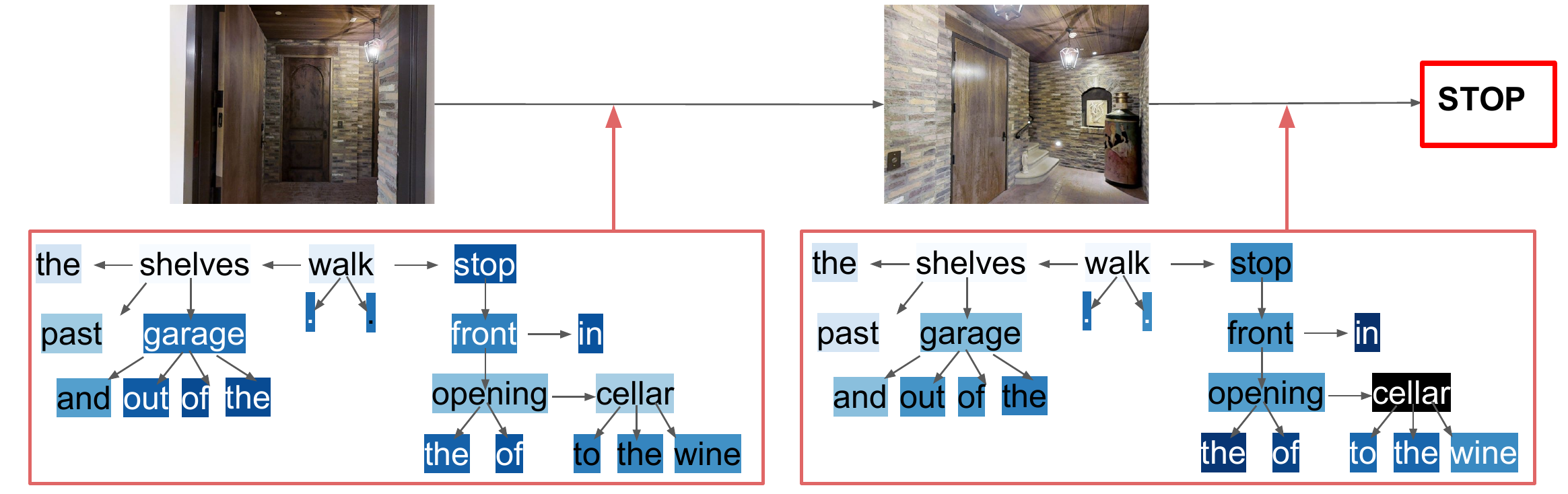}
\caption{The weights for the grounded instructions for our syntax-aware model.}
\label{fig3}
\end{figure*}

\begin{figure}[t]
\centering
\includegraphics[width=1.0\columnwidth]{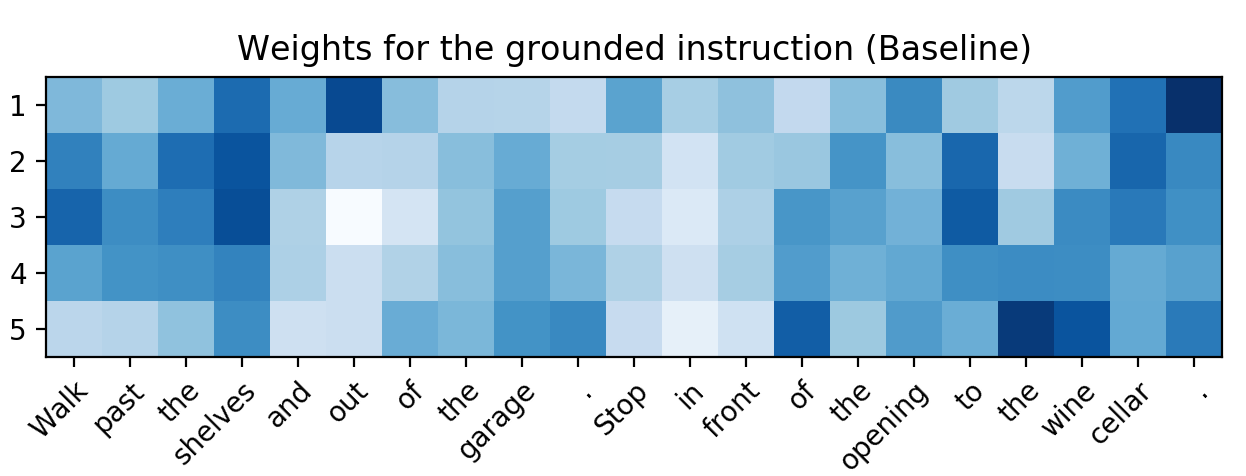}
\vspace{-15pt}
\caption{The weights for the grounded instructions for baseline model.}
\label{fig4}
\end{figure}

\section{Experimental Setup}

\subsection{Datasets}
We evaluate our agent on Room-to-Room (R2R) dataset \citep{anderson2018vision} and Room-Across-Room (RxR) dataset \citep{ku2020room}. Both datasets are built on Matterport3D simulator \citep{anderson2018vision}.
The R2R dataset contains 21567 human-annotated instructions with an average instruction length of 29. The dataset is divided into training set, seen validation set, unseen validation set, and test set. 
The RxR dataset is an extension of the R2R dataset. The instructions are longer (with an average instruction length of 78), and the instructions are in three languages (i.e., English, Hindi, and Telugu). The RxR dataset follows the same division as the R2R dataset. Details can be found in the Appendix.

\subsection{Evaluation Metrics}

To evaluate the performance of our model, we use the following evaluation metrics: Success Rate (SR), Success Rate Weighted by Path Length (SPL) \citep{anderson2018evaluation}, normalized Dynamic Time Warping (nDTW) \citep{magalhaes2019effective}, success rate weighted by Dynamic Time Warping \citep{magalhaes2019effective} and Coverage weighted by Length Score (CLS). Detailed description for each metric can be found in Appendix.

\section{Results and Analysis}

\subsection{Room-to-Room Dataset}
We compare our agent with the baseline agent \citep{tan-etal-2019-learning}\footnote{The baseline is the re-implementation of their model without back translation based on code: https://github.com/airsplay/R2R-EnvDrop} on the R2R test leaderboard. Our syntax-aware agent achieves 47.8\% in success rate and 0.45 in SPL, improving the baseline model by 2.1\% in success rate and 2\% in SPL. 

Besides, as shown in Table \ref{table2}, our syntax-aware agent achieves 3.7\% improvement in success rate over the baseline model in validation unseen environment, indicating that explicitly incorporating syntax information can better guide the agent to ground to visual scenes.

We further experiment on R2R dataset to see whether the increase in performance comes from more model parameters. Compared with the model that uses a 2-layer LSTM (+LSTM), we can see that our syntax-aware model still achieves 2.6\% increase in success rate in validation unseen environment. This result validates the effectiveness of incorporating syntax information. 

\subsection{Room-Across-Room Dataset}
We first compare our agent with the baseline agent in RxR paper \cite{ku2020room} on the RxR test leaderboard. Our syntax-aware agent (``SAA (mono)'' on the leaderboard) outperforms the baseline in all metrics, improving the nDTW score by 5.73\% and success rate by 9.98\%.
Moreover, We compare our agent with the baseline on RxR validation set. As shown in Table \ref{table3}, in all three languages, our syntax-aware agent outperforms the baseline agent in all metrics on validation unseen set. Specifically, our model gets 2.6\% improvement in Hindi instructions in terms of success rate. For English and Telugu instructions, our model gets smaller improvement -- 0.7\% and 0.1\% respectively. One reason for these results could be the correlation with the quality of the dependency parser in that language. Besides, compared with the baseline in \citet{ku2020room}, our agent achieves the new state-of-the-art on RxR dataset (at the time of submission).

\subsection{Qualitative Analysis}
As shown in Fig \ref{fig3} and Fig \ref{fig4}, we illustrate a qualitative example to show that our agent learns better alignment between instruction and visual scenes. As shown in Fig \ref{fig4}, given instruction \textit{``Walk past the shelves and out of the garage. Stop in front of the opening to the wine cellar."}, the baseline agent tends to focus on the same phrases (e.g., ``past the shelves'', ``wine cellar'') during navigation. This suggests that the baseline agent is not able to learn a good alignment between the instruction and the current visual scene. However, as shown in Figure \ref{fig3}, our dependency-based agent can successfully identify the correct parts of the instruction that are correlated with the current visual scenes, and picks the next action with fidelity. At the beginning of navigation, our agent focuses on the first sub-instruction ``walk past the shelves''. Then ``out of the garage" gradually becomes the most important phrase, indicating that the agent should go out of the garage after passing the shelves. When the agent sees the wine sculpture, it infers that the opening is for wine cellar and stops in front of the opening near the wine sculpture. 

\subsection{Implementation Variants}
Since the goal of our paper is to explore the role of syntax information in vision-language navigation, we try several implementation variants to include syntax information. First, we try to use mean-pooling instead of Tree-LSTM to encode the dependency tree structure. This implementation variant decreases the performance by around 3\% in terms of success rate on validation unseen environments. Besides, we explore whether syntax information from a constituency tree can also help with the instruction following and navigation. Similar to how we incorporate dependency tree information, we use a Tree-LSTM to encode the constituency tree information. However, the performance decreases around 2\% in terms of success rate in validation unseen environments, indicating that syntax information extracted from dependency tree is more beneficial for the vision-and-language navigation task.

\section{Conclusion}
In this paper, we presented a tree-based encoder module that incorporates phrase-level information from parse trees. We demonstrated that syntax information can help the agent learn a better alignment between instruction and visual scenes, and generalize better to unseen environments. Our experiments on Room-to-Room dataset and Room-Across-Room dataset both suggest that incorporating syntax information encoded by our tree-based encoder module can significantly improve the performance over baseline VLN models.

\section{Ethical Considerations}
Vision-Language Navigation is the task that requires an agent to navigate through a 3D environment based on given natural language instructions. An agent that can interact with the environment based on instructions can be used in many real-world applications, for example, a home service robot can bring things to the owner based on instruction, making people's life easier. However, when deployed in the real world, even if the agent can navigate successfully based on the instruction, it might still need further human assistance to keep working successfully (e.g., a home-cleaning robot might be stuck in the corner of the room and cannot get out by itself).

The performance on the validation set with the unseen environment is much lower than the seen environment. Change in environment will significantly influence the performance of the agent. When the agent is deployed in an unseen environment, it will have a higher probability of failure, wasting energy and time. A further pre-exploration of the environment will be needed for better performance of the agent when deployed to real-world applications. Moreover, our agent relies on the quality of the dependency parser to some extent. Though we achieve improvement in all three languages when using the dependency tree information, the agent in Hindi and English benefit most from the syntax information because of the best available parser for these languages.

\section{Acknowledgement}
We thank the reviewers for their helpful discussions. This work was supported by ARO-YIP Award W911NF-18-1-0336, DARPA MCS Grant N66001-19-2-4031, a Google Focused Research Award, and a Bloomberg Data Science Ph.D. Fellowship. The views, opinions, and/or findings contained in this article are those of the authors and not of the funding agency.

\bibliography{anthology,custom}
\bibliographystyle{acl_natbib}

\appendix

\section{Appendix}

\subsection{Problem Setup}
Vision-Language navigation requires an agent to navigate through a 3D environment to a target location based on a given natural language instruction. Formally, the natural language instruction is a sequence of words $\{w_i\}_{i=1}^{l}$, where $l$ is the length of the instruction and $w_i$ is $ith$ word in the sequence. At each time step, the agent perceives a panoramic view of the current viewpoint. Besides, the agent has access to a set of navigable locations $\{l_{t,k}\}_{k=1}^{K}$, where $K$ is the total number of reachable locations from the current viewpoint. The agent needs to select an action $a_t$ from the list of navigable viewpoints $\{l_{t,k}\}_{k=1}^{K}$ based on the given instruction, navigation history and current panoramic views. If the viewpoint selected from the list is the same as the current viewpoint, the agent predicts a ``STOP'' action.

\subsection{Model Details}
\textbf{Visual Encoder.}
Same as previous work in VLN, at time step $t$, we discretize the panoramic view into 36 single views $\{o_{t,p}\}_{p=1}^{36}$. Each single view is a RGB image, annotated with its angles of heading and elevation $(\theta_{t,p}, \phi_{t,p})$. Each RGB image is encoded with a pre-trained ResNet-152 \citep{he2016deep} on ImageNet \citep{ILSVRC15}. A four dimension orientation feature $(cos\theta_{t,p}, sin\theta_{t,p}, cos\phi_{t,p}, sin\phi_{t,p})$ is concatenated with the ResNet feature to form the final representation for the view $p$ of a panoramic $\{f_{t,p}\}_{p=1}^{36}$. Similarly, we get the visual representation $\{g_{t,k}\}_{k=1}^K$ for $K$ navigable locations $\{l_{t,k}\}_{k=1}^K$ at time step $t$.

\subsection{Training}
We use a mixture of imitation learning and reinforcement learning to train the agent. 

\textbf{Imitation Learning.} 
During training, instead of navigating to the predicted action at each time step, teacher-forcing is used to determine which navigable viewpoint to pick. Given the shortest path between the start point and target point, at each time step $t$, the agent tries to imitate the teacher action $a_t^\star$ by minimizing the negative log probability:
\begin{align}
    L_{IL} &= \sum_t{-{a_t^\star}logp_t}
\end{align}
\textbf{Reinforcement Learning.}
We combine imitation learning with reinforcement learning to learn a more generalizable agent. Since the teacher path is the shortest path between the start point and the target point, there is no guarantee that the teacher path is the same as indicated by the given instruction. Thus, reinforcement learning is applied for better instruction following and state exploration. At each time step $t$, the agent samples an action $a_t$ from the predicted distribution $p_t(a_{t})$. At each time step, if the agent moves closer to the target viewpoint, a positive reward +1 is given, otherwise the agent receives a negative reward -1. When the agent predicts the ``STOP" action, the agent will receive a +3/-3 reward based on whether the agent is within 3m from the target viewpoint. We use Actor-Critic \citep{mnih2016asynchronous} to train the agent. The loss of reinforcement learning is:
\begin{align}
    L_{RL} &= \sum_t{(R_t - R_{b_t})logp_t(a_t) + \eta H(p_t(a_t))}
\end{align}
where $R_t$ is the discounted future cumulative rewards at time step $t$, $R_{b_t}$ is the expected cumulative rewards (baseline) approximated by the value function $V$, $H(p_t(a_t))$ is the entropy term for regularization. Specifically, 
\begin{align}
    R_t &= r_t + \sum_{i=1}^{T-t}\gamma^{i}r_{t+i} \\
    R_{b_t} &= V(h_t) = W_{v_2}\sigma(W_{v_1}h_t)
\end{align}
where $\sigma$ is the ReLU activation function, and $r_t$ is the immediate reward we defined earlier. The value function is trained with L2 loss:
\begin{align}
    L_{V} &= \frac{1}{2}(R_t - R_{b_t})^2
\end{align}
We optimize a mixture loss of imitation learning and reinforcement learning:
\begin{align}
    L_{MIX} &= (L_{RL} + L_{V}) + \lambda L_{IL}
\end{align}

\subsection{Dataset}
\textbf{R2R Dataset.} The R2R dataset contains 21567 human annotated instructions with an average instruction lengths of 29. The training set contains 14025 instructions in 61 environments. The seen validation set contains 1020 instructions in the same 61 environments as the training set. The unseen validation set contains 2349 instructions in 11 environment which is not included in the training set. The test set contains 4173 instructions in 18 environments. 

\textbf{RxR Dataset.} The RxR dataset is an extension to the R2R dataset, where the instructions are longer and in languages other than English (Hindi and Telugu). It contains 126069 instructions with an average instruction length of 78. Besides, different from R2R that only contains guide path (i.e., the shortest path between the start point and the target point), RxR pairs each guide path with a human-annotated follower path (i.e., the path that human generates following the instruction). We only use the guide path to train the agent in this paper. The seen and unseen environment split is the same as in the R2R dataset. There are 16522 paths in total, and each path is annotated in 3 languages. The training set contains 11089 paths, the seen validation set contains 1232 paths, the unseen validation contains 1517 paths, and the test set contains 2684 paths. 

\subsection{Evaluation Metrics}
To evaluate the performance of our model, we use the following evaluation metrics: (1) Success Rate (SR): If the agent stops less than 3m from the target location, we consider the navigation as a success. (2) Success Rate Weighted by Path Length (SPL) \citep{anderson2018evaluation}: This metric penalizes long trajectories (e.g., find the target using beam search over the environment graph). (3) normalized Dynamic Time Warping (nDTW) \citep{magalhaes2019effective}: This metric penalizes deviations from the reference path. (4) success rate weighted by Dynamic Time Warping (sDTW) \citep{magalhaes2019effective}: This metric constraints nDTW only to successful navigation and considers path fidelity and agent success. (5) Coverage weighted by Length Score (CLS) \citep{jain-etal-2019-stay}: Similar to nDTW, this metric also encourages path fidelity.

\subsection{Implementation Details}
We generate the dependency parse tree for English using Stanford CoreNLP \citep{manning2014stanford}. We use Stanza Toolkit \citep{qi-etal-2020-stanza} to generate the dependency parse tree for Hindi and Telugu. For both baseline model and our syntax-aware model, the learned word embedding size is 256 and the dimension of the action embedding is 128. We set the hidden size for the bi-directional LSTM to be 256. For syntax-aware model, the hidden size for the bi-LSTM and Tree-LSTM in the language encoder is 512. We use ResNet-152 \citep{he2016deep} to extract the 2048 dimensional image feature. We use RMSProp \citep{hinton2012neural} as the optimizer with learning rate 1e-4 and batch size 64. The weight $\lambda$ we use to combine imitation learning loss and reinforcement learning loss is set to be 0.2. In reinforcement learning, the discount factor $\gamma$ is 0.9 and the entropy weight $\eta$ is 0.01. During training, we set the max action length to be 35 for R2R dataset and 70 for RxR dataset. We train both agents on R2R for 80,000 iterations. We train both agents on RxR for 200,000 iterations. The baseline model contains approximate 6 million parameters. Our syntax-aware model contain approximate 8 million parameters.

\end{document}